\documentclass[letterpaper, 10 pt, conference]{ieeeconf}  
\IEEEoverridecommandlockouts                              
\pdfoutput=1

\usepackage[table]{xcolor}
\usepackage{graphicx} 
\usepackage[caption=false]{subfig} 
\graphicspath{{images/}} 
\DeclareGraphicsExtensions{.pdf,.jpg,.png} 
\usepackage{amsmath} 
\usepackage{amsfonts} 
\usepackage{tikz} 
\usetikzlibrary{shapes,arrows,fit,calc,positioning,automata}
\tikzset{>=latex}
\usepackage{dblfloatfix} 
\usepackage{cite}
\usepackage{color} 
\usepackage{multirow} 
\usepackage{hhline} 
\usepackage{hyperref} 
\usepackage{tabularx} 
\usepackage{epstopdf} 
\usepackage{enumerate} 
\usepackage{multicol} 
\usepackage{siunitx} 
\usepackage{mathtools} 
\usepackage{cuted} 
\usepackage[ruled,vlined]{algorithm2e} 
\usepackage{cleveref} 

\newtheorem{remark}{Remark}

\newcommand{\argmin}[1]{\underset{#1}{\operatorname{arg}\,\operatorname{min}}\;}

\title{\LARGE \bf
Online Deep Learning for Improved Trajectory Tracking of \\ Unmanned Aerial Vehicles Using Expert Knowledge
}

\author{Andriy Sarabakha$^{1}$ and Erdal Kayacan$^{2}$
\thanks{$^{1}$Andriy Sarabakha is with School of Mechanical and Aerospace Engineering, Nanyang Technological University, 50 Nanyang Avenue, Singapore, 639798. {\tt\small andriy001@e.ntu.edu.sg}}%
\thanks{$^{2}$Erdal Kayacan is with Department of Engineering, Aarhus University, Aabogade 34, Aarhus N, 8200, Denmark. {\tt\small erdal@eng.au.dk}}%
}

\begin{document}

\maketitle
\thispagestyle{empty}
\pagestyle{empty}

\begin{abstract}

This work presents an online learning-based control method for improved trajectory tracking of unmanned aerial vehicles using both deep learning and expert knowledge. The proposed method does not require the exact model of the system to be controlled, and it is robust against variations in system dynamics as well as operational uncertainties. The learning is divided into two phases: offline (pre-)training and online (post-)training. In the former, a conventional controller performs a set of trajectories and, based on the input-output dataset, the deep neural network (DNN)-based controller is trained. In the latter, the trained DNN, which mimics the conventional controller, controls the system. Unlike the existing papers in the literature, the network is still being trained for different sets of trajectories which are not used in the training phase of DNN. Thanks to the rule-base, which contains the expert knowledge, the proposed framework learns the system dynamics and operational uncertainties in real-time. The experimental results show that the proposed online learning-based approach gives better trajectory tracking performance when compared to the only offline trained network.

\end{abstract}

\section{INTRODUCTION}
\label{sec:introduction}

Research activities to develop increased autonomy in unmanned aerial vehicles (UAVs) have taken a centre stage in the recent years due to their usefulness in providing cost-effective solutions to dangerous, dirty and dull tasks, such as aerial grasping \cite{Loianno2018RAL}, emergency evacuation \cite{Sarabakha2016CDC} and building inspection \cite{Teixeira2017ICRA}. In these applications, it is crucial for UAVs to be able to fly autonomously in uncertain environments with variations in operating conditions \cite{Sanket2018RAL}.
Therefore, in such conditions, adaptability is a must rather than a choice.  

Given the ability of artificial neural networks (ANNs) to generalise knowledge from training samples, an ANN-based controller can be used to control nonlinear dynamic systems \cite{Kayacan2013TC}. On the other hand, deep neural networks (DNNs) can approximate non-linear functions with exponentially lower number of training parameters and higher sample complexity when compared to ANNs \cite{LeCun2015Nature}. Therefore, DNNs propose a novel approach to enhance the control strategies \cite{Zhou2018RAL}. 

In the literature, ANNs have successfully been integrated with control system design to improve tracking performance in uncertain environments \cite{Emran2017SMC}. In \cite{Bansal2016CDC}, the unknown part of the dynamical model of a quadcopter is modelled by DNN. In \cite{Nivison2017ACC}, DNN is used for direct inverse control of the quadrotor in simulation. In \cite{Punjani2015ICRA} and \cite{Mohajerin2014SMC}, DNNs are used to learn the dynamics of helicopter and multicopter, respectively. In \cite{Li2017ICRA}, DNN pre-cascaded module is used to improve the performance of UAV in tracking arbitrary hand-drawn trajectory. However, in all these works, DNNs are trained offline and, then, used online without further learning. In other words, while the dynamics are learnt in the training phase, the controller is not updated in the testing phase -- DNN simply mimics the conventional controller -- and the operational uncertainties are no longer learnt.

\begin{figure}[!t]
\centering
\includegraphics[width=1\columnwidth]{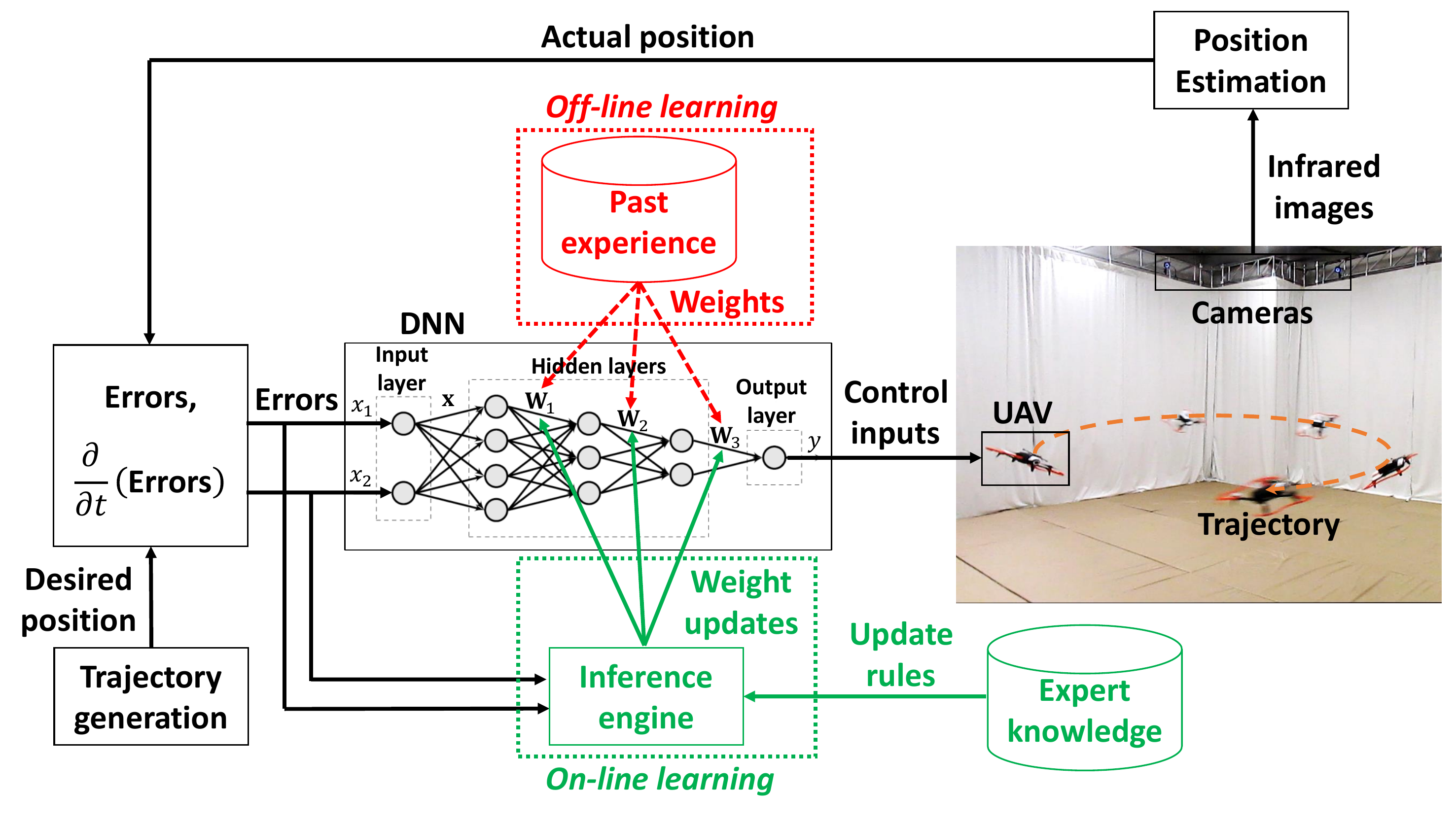}
\caption{Illustration of the proposed online learning-based control scheme. During the offline learning phase, a conventional controller performs a set of trajectories and the input-output dataset is used for the offline training of DNN. During the online learning phase, DNN controls the UAV and, by using the expert knowledge, improves further the performance in real-time.
}
\label{fig:architecture}
\end{figure}

Unlike the traditional use of DNNs in literature, in this work, we propose an online DNN-based approach for improving trajectory tracking performance of UAVs. After an offline pre-training phase with past flight data, a DNN-based controller is used in real-time to control the UAV. Without any prior knowledge of the system, besides the training data, the proposed approach shows its capability to reduce the trajectory tracking error online by compensating for internal uncertainties and external disturbances. Moreover, it is shown that the DNN module is computationally suitable for real-time operations and adequate for arbitrary trajectory, making it applicable to the real-world tasks. Furthermore, the proposed approach employs the expert knowledge for the online training. The overall control architecture and its training process are depicted in Fig.~\ref{fig:architecture}.

\begin{table*}[!b]\begin{normalsize}
\begin{equation*}
\begin{split}
f(\mathbf{x}) = \begin{bmatrix}u & v & w & p + s_{\phi} t_{\theta} q + c_{\phi} t_{\theta} r & c_{\phi} q - s_{\phi} r & \frac{s_{\phi}}{c_{\theta}} q + \frac{c_{\phi}}{c_{\theta}} r & 0 & 0 & g & \frac{I_y - I_z}{I_x} q r & \frac{I_z - I_x}{I_y} p r & \frac{I_x - I_y}{I_z} p q \end{bmatrix}^{T}, \\
g(\mathbf{x}) = \begin{bmatrix} 0 & 0 & 0 & 0 & 0 & 0 & -\frac{1}{m} \left( c_{\phi} c_{\psi} s_{\theta} + s_{\phi} s_{\psi} \right) & -\frac{1}{m} \left( c_{\phi} s_{\psi} s_{\theta} - c_{\psi} s_{\phi} \right) & -\frac{1}{m} c_{\phi} c_{\theta} & 0 & 0 & 0 \\ 0 & 0 & 0 & 0 & 0 & 0 & 0 & 0 & 0 & \frac{1}{I_x} & 0 & 0 \\ 0 & 0 & 0 & 0 & 0 & 0 & 0 & 0 & 0 & 0 & \frac{1}{I_y} & 0 \\ 0 & 0 & 0 & 0 & 0 & 0 & 0 & 0 & 0 & 0 & 0 & \frac{1}{I_z} \end{bmatrix}^{T},
\end{split}
\end{equation*}
\end{normalsize}\end{table*}

This work is organised as follows. The problem is formulated in Section~\ref{sec:problem}. Section~\ref{sec:dnn} introduces the proposed approach. Then, Section~\ref{sec:setup} presents the experimental setup. Section~\ref{sec:results} provides real-time experiments with quadcopter UAV, to validate the proposed method. Finally, Section~\ref{sec:conclusions} summarises this work with conclusions and future work.

\section{PROBLEM FORMULATION}
\label{sec:problem}

In this work, we consider a problem of designing a learning feedback control algorithm for a dynamical system, such as UAV. Our objective is to learn a control strategy of the system to achieve a high-accuracy tracking. To describe the problem, we introduce the dynamical model of UAV first.


\subsection{Dynamical Model of Unmanned Aerial Vehicle}

The world-fixed reference frame is $\mathcal{F}_W = \{\vec{\mathbf{x}}_W, \vec{\mathbf{y}}_W, \vec{\mathbf{z}}_W\}$ and the body frame is $\mathcal{F}_B = \{\vec{\mathbf{x}}_B, \vec{\mathbf{y}}_B, \vec{\mathbf{z}}_B\}$. 
The absolute position of UAV $\mathbf{p} = \begin{bmatrix} x & y & z \end{bmatrix}^T$ is given by three Cartesian coordinates at its center of gravity in $\mathcal{F}_W$, and its attitude $\mathbf{o} = \begin{bmatrix} \phi & \theta & \psi \end{bmatrix}^T$ is given by three Euler angles. The rotation matrix from $\mathcal{F}_B$ to $\mathcal{F}_W$ is given by the combination of three single rotation matrices around $\phi$, $\theta$ and $\psi$.
The time derivative of the position gives the linear velocity $\mathbf{v} = \begin{bmatrix}\dot{x} & \dot{y} & \dot{z}\end{bmatrix}^T = \begin{bmatrix} v_x & v_y & v_z \end{bmatrix}^T$ of UAV expressed in $\mathcal{F}_W$. Equivalently, the time derivative of the attitude $\boldsymbol{\omega} = \begin{bmatrix}\dot{\phi} & \dot{\theta} & \dot{\psi}\end{bmatrix}^T$ gives the angular velocity in $\mathcal{F}_W$ and $\boldsymbol{\omega}_B = \begin{bmatrix} p & q & r \end{bmatrix}^{T}$ is the angular velocity in $\mathcal{F}_B$.

The vector of control inputs $\mathbf{u}$ is chosen as:
\begin{equation}
\mathbf{u} = \begin{bmatrix} T & \tau_{\phi} & \tau_{\theta} & \tau_{\psi} \end{bmatrix}^T,
\label{eq:inputs}
\end{equation}
where $T$ is the total thrust along $\vec{\mathbf{z}}_B$, whereas $\tau_{\phi}$, $\tau_{\theta}$ and $\tau_{\psi}$ are moments around $\vec{\mathbf{x}}_B$, $\vec{\mathbf{y}}_B$ and $\vec{\mathbf{z}}_B$, respectively. Finally, the dynamical model of UAV is given as in \cite{Mahony2012RAM}:
\begin{equation}
\begin{cases}
\dot{x} = v_x & \dot{u} = \frac{1}{m} \left( c_{\phi} c_{\psi} s_{\theta} + s_{\phi} s_{\psi} \right) T \\
\dot{y} = v_y & \dot{v} = \frac{1}{m} \left( c_{\phi} s_{\psi} s_{\theta} - c_{\psi} s_{\phi} \right) T \\
\dot{z} = v_z & \dot{w} = -g + \frac{1}{m} c_{\phi} c_{\theta} T \\
\dot{\phi} = p + s_{\phi} t_{\theta} q + c_{\phi} t_{\theta} r & \dot{p} =  \frac{I_y - I_z}{I_x} q r + \frac{1}{I_x} \tau_{\phi} \\
\dot{\theta} = c_{\phi} q - s_{\phi} r & \dot{q} =  \frac{I_z - I_x}{I_y} p r + \frac{1}{I_y} \tau_{\theta} \\
\dot{\psi} = \frac{s_{\phi}}{c_{\theta}} q + \frac{c_{\phi}}{c_{\theta}} r & \dot{r} =  \frac{I_x - I_y}{I_z} p q + \frac{1}{I_z} \tau_{\psi},
\end{cases}
\label{eq:model}
\end{equation}
where $m$ is the mass of UAV, $g$ is the gravity acceleration constant, $\mathbf{I} = \mathrm{diag}(I_x, I_y, I_z)$ is the inertia matrix, $c_{\star}$, $s_{\star}$ and $t_{\star}$ denote $\cos{(\star)}$, $\sin{(\star)}$ and $\tan{(\star)}$, respectively.

\begin{remark}
The dynamical system in (\ref{eq:model}) is nonlinear, coupled and underactuated. Therefore, an advanced controller is required.
\end{remark}

The system in (\ref{eq:model}) can be written in a general form as:

\begin{equation}
\begin{cases}
\dot{\mathbf{x}} = f(\mathbf{x}) + g(\mathbf{x}) \mathbf{u} + d\\
\mathbf{y} = h(\mathbf{x}),
\end{cases}
\label{eq:nonlinear_system}
\end{equation}
where $\mathbf{x} = \begin{bmatrix} x & y & z & \phi & \theta & \psi & u & v & w & p & q & r \end{bmatrix}^{T}$,

\noindent $d$ is the disturbance term, $h(\mathbf{x}) = \begin{bmatrix} x & y & z & \phi & \theta & \psi \end{bmatrix}^{T}$ and $\mathbf{u}$ is defined in (\ref{eq:inputs}).

\subsection{Problem Description}

If a precise model of the system exists, then the inversion of the system can be computed. Let $h  \circ f$ denote the composition of functions $h$ and $f$; while $f^i$ denote the $i$-th composition of function $f$, i.e., $f^0(\mathbf{x}) = \mathbf{x}$ and $f^i(\mathbf{x}) = f^{i - 1} \circ f(\mathbf{x}) \quad \forall i \in \mathbb{N}^+$ \cite{Sun2001DC}. Let $n$ define the dimension of the system's input, i.e., $\mathbf{u} \in \mathbb{R}^n$, and let $\mathbf{r}$ define the vector of relative degrees of the system, s.t. $\argmin{\mathbf{r}_i} \frac{\partial}{\partial \mathbf{u}_i} \left( h \circ f^{\mathbf{r}_i - 1} \circ \left( f(\mathbf{x}) + g(\mathbf{x}) \mathbf{u} \right) \right) \neq 0 \quad \forall i \in [1, n]$. Then, the input and the output of the system are related by
\begin{equation}
\mathbf{y}_{k + \mathbf{r}_i} = h \circ f^{\mathbf{r}_i - 1} \circ \left( f(\mathbf{x}_k) + g(\mathbf{x}_k) \mathbf{u}_k \right).
\label{eq:input_output}
\end{equation}
If $\mathbf{y}$ is affine in $\mathbf{u}$, then (\ref{eq:input_output}) becomes
\begin{equation}
\mathbf{y}_{k + \mathbf{r}_i} = F(\mathbf{x}_k) + G(\mathbf{x}_k) \mathbf{u}_k,
\label{eq:input_output_affine}
\end{equation}
where $F_i(\mathbf{x}_k) = h \circ f^{\mathbf{r}_i}(\mathbf{x}_k)$ and $G_i(\mathbf{x}_k) = \frac{\partial}{\partial \mathbf{u}_{k, i}} \left( h \circ f^{\mathbf{r}_i - 1} \circ \left( f(\mathbf{x}_k) + g(\mathbf{x}_k) \mathbf{u}_k \right) \right)$ are the decoupling matrices. Finally, the control law at time $k$ to track the desired output of the system $\mathbf{y}^*$ can be written as in \cite{Zhou2017CDC}:
\begin{equation}
\mathbf{u}_{k, i} = \left[ G(\mathbf{x}_k) \right]^{-1} \left( \mathbf{y}^*_{k + \mathbf{r}_i} - F(\mathbf{x}_k) \right).
\label{eq:system_inversion}
\end{equation}

However, in a real system, the system's parameters might be unknown and difficult to estimate, e.g., moments of inertia. What is more, these parameters might change during the operation of the system, e.g., mass. Moreover, it is not always possible to predict the external disturbance term. Therefore, an adaptive controller which can learn online is required. Our objective is to learn the control of the system by only looking at the performance of the system, i.e., in our case, the tracking error:
\begin{equation}
\mathbf{e}_k = \mathbf{y}^*_k - \mathbf{y}_k,
\label{eq:error}
\end{equation}
and its time derivative:
\begin{equation}
\dot{\mathbf{e}}_k = \dot{\mathbf{y}}^*_k - \dot{\mathbf{y}}_k.
\label{eq:d_error}
\end{equation}
Thus, $\mathbf{y}_k$ and $\dot{\mathbf{y}}_k$ is the only required information about the system. 

\section{METHODOLOGY}
\label{sec:dnn}

By their nature, DNNs are distinguished from more common single-hidden-layer ANNs by their depth. The neurons are organised in input, multiple-hidden and output layers. In DNN, like in classical ANNs, the weights are modified using a learning process governed by the training rules.

\noindent \hrulefill

\subsection{Offline Pre-Training}

\begin{figure}[!b]
\centering
\subfloat[Block diagram of the offline pre-training of DNN by PID.]{\includegraphics[width=0.48\columnwidth]{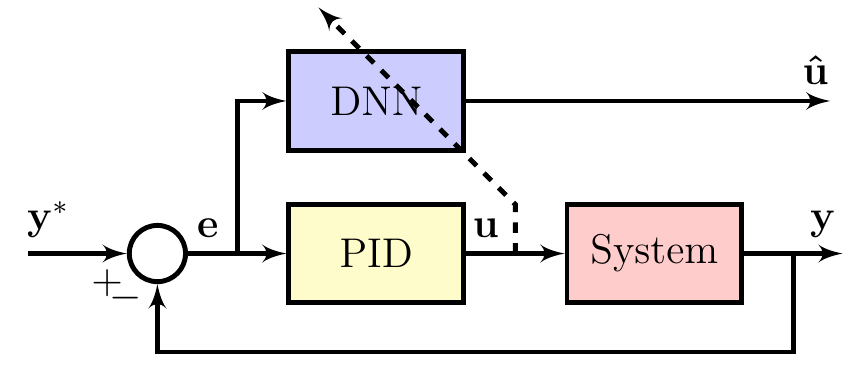}
\label{fig:pretraining}} \hfil
\subfloat[Block diagram of the online post-training of DNN by FLS.]{\includegraphics[width=0.48\columnwidth]{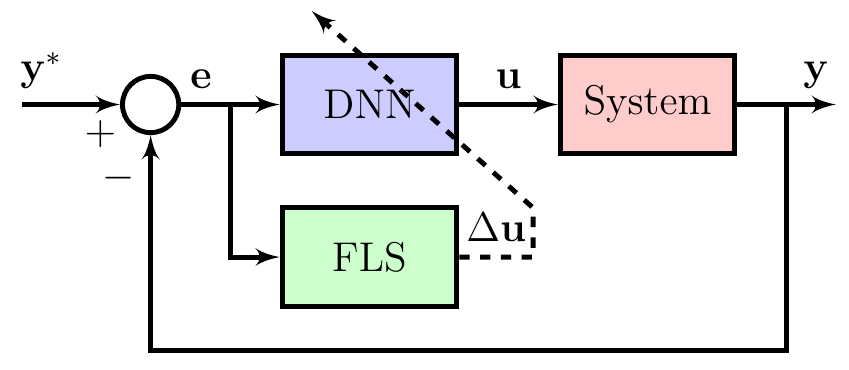}
\label{fig:training}}
\caption{Block diagrams of two control paradigms: offline pre-training and online post-training.}
\label{fig:blocks}
\end{figure}

During the offline pre-training phase, a supervised learning approach is used, in which a feed-forward DNN learns to control the system from a conventional controller -- proportional-integral-derivative (PID) controller, in our case. In this control scheme, shown in Fig.~\ref{fig:pretraining}, PID controller controls the system alone. Hence, it is utilized as an ordinary feedback controller to ensure the global asymptotic stability of the system and provide labelled training samples for DNN. The training of DNN requires the availability of a large number of labelled training samples. Each labelled training sample consists of an input and expected output pair $<\{ \mathbf{e}_k, \mathbf{\dot{e}}_k \}, \{ \mathbf{u}_k \}>$. The training of DNN involves back-propagation to minimize the loss over all training examples. After the training, DNN can approximate the mapping from the training inputs to the outputs. The pseudo-code of offline pre-training is provided in Algorithm~\ref{alg:offline}.

\subsection{Online Training}

During the online training phase, DNN controls the system, and, at the same time, learns how to improve the control performances. Since DNN training requires supervised learning, another process has to provide a feedback about its performances. In our case, fuzzy logic system (FLS) is used to provide this information. By definition, FLS incorporates the expert knowledge in form of rules and uses this knowledge to provide some useful information \cite{Mendel2017Book}. The control structure for online training is illustrated in Fig.~\ref{fig:training}.

\begin{figure}[!b]
\centering
\includegraphics[width=0.9\columnwidth]{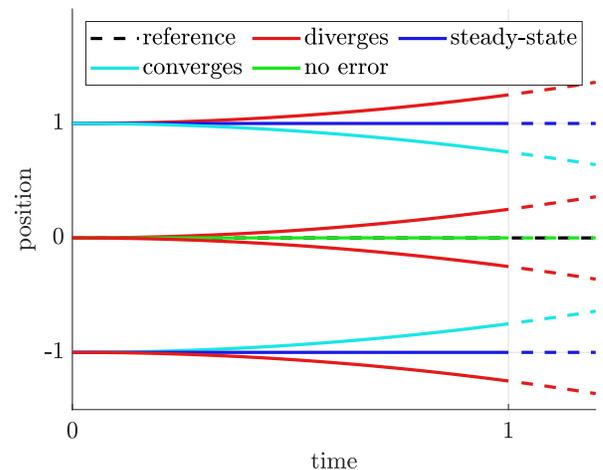}
\caption{Possible evolution of the tracking error in a dynamical system. The system can diverge (red curves), converge (cyan curves), it can have a steady-state error (blue lines), or the error can be zero (green line).}
\label{fig:evolution}
\end{figure}

In our approach, FLS observes the behaviour of the system controlled by DNN, and, depending on the situation, corrects the action of DNN. The possible evolutions of the error are depicted in Fig.~\ref{fig:evolution}. If the error is positive, i.e, $e_i > 0$, and its time derivative is also positive, i.e., $\dot{e}_i > 0$, then the system diverges (top red curve in Fig.~\ref{fig:evolution}). In this case, FLS will force DNN to decrease the control signal $u_i$ significantly to stabilize the system, i.e., $\Delta u_i \ll 0$. In another possible case, if the error is negative, i.e., $e_i < 0$, and its time derivative is zero, i.e., $\dot{e}_i = 0$, then the error is steady (bottom blue line in Fig.~\ref{fig:evolution}). In this case, DNN falls down in a local minimum and FLS will give a small positive perturbation, i.e., $\Delta u_i > 0$. Finally, if the error is zero, i.e., $e_i = 0$, and its time derivative is also zero, i.e., $\dot{e}_i = 0$, then, this is the optimal case (green line in Fig.~\ref{fig:evolution}) and no action has to be taken, i.e., $\Delta u_i = 0$.

\begin{algorithm}[!t]
	\SetAlgoLined
	\DontPrintSemicolon
	\KwIn{-}
	\KwOut{Pre-trained DNN$_0$}
	\Begin{			
		\While{$k < $ MaxSamples}{
			Get $\mathbf{y}_k$, $\mathbf{y}^*_k$, $\dot{\mathbf{y}}_k$, $\dot{\mathbf{y}}^*_k$ and $\mathbf{u}_k$\;
			$\mathbf{e}_k \gets \mathbf{y}^*_k - \mathbf{y}_k$ by using (\ref{eq:error})\;
			$\dot{\mathbf{e}}_k \gets \dot{\mathbf{y}}^*_k - \dot{\mathbf{y}}_k$ by using (\ref{eq:d_error})\;
			Collect $<\{ \mathbf{e}_k, \dot{\mathbf{e}}_k \}, \{ \mathbf{u}_k \}>$\;
		}
		DNN$_0 \gets $ ConstructNetworkLayers()\;
		$\mathbf{w} \gets $ InitializeWeights()\;
		Train DNN$_0$ on $<\{ \mathbf{e}, \dot{\mathbf{e}} \}, \{ \mathbf{u} \}>$
	}
	\label{alg:offline}
	\caption{Offline pre-training of DNN.}
\end{algorithm}

These empirical rules can be formally described by a Mamdani FLS with triangular membership functions to represent the fuzzy sets. The rules for each possible case are summarized by the rule-base in Table~\ref{tab:rulebase}. The inputs to the FLC are selected to be the tracking error and its time derivative, i.e., $e_i$ and $\dot{e}_i$; while the output is the correction signal, i.e., $\Delta u_i$. The input is represented by three fuzzy sets: negative, zero and positive; while the output can belong to five fuzzy sets: big decrease, small decrease, no changes, small increase and big increase.

However, FLS requires operations among fuzzy sets which are time-consuming. Therefore, by using a similar approach to the one described in \cite{Sarabakha2018TIE}, a fuzzy mapping which represents the FLS in Table~\ref{tab:rulebase} can be generated for a general multidimensional case:
\begin{equation}
\Delta \mathbf{u}_k = -\alpha \left( \frac{1}{2} \mathbf{e}_k + \dot{\mathbf{e}}_k - \frac{1}{2} \text{abs}(\mathbf{e}_k) \odot \dot{\mathbf{e}}_k \right),
\label{eq:fm}
\end{equation}
where $\odot$ denotes Hadamard product and $\alpha$ is the adaptation rate. The fuzzy mapping reduces significantly the computation time which makes this approach suitable for real-time systems \cite{Sarabakha2017FUZZIEEE}. The pseudo-code of online training is provided in Algorithm~\ref{alg:online}.


\begin{table}[!t]
\centering
\caption{Rule-base for the updates of $u_i$.}
\begin{tabular}{|l||l|l|l|} \hline
\multicolumn{1}{|c||}{\multirow{2}{*}{$e_i$}} 	& \multicolumn{3}{c|}{$\dot{e}_i$} 																												\\ \hhline{~---}
 												& Negative 					& Zero 						& Positive							\\ \hline \hline
Negative 										& \textbf{Big decrease} 	& \textbf{Small decrease} 	& \textbf{No changes} 	\\ \hline
Zero  											& \textbf{Big decrease} 	& \textbf{No changes} 		& \textbf{Big increase} 	\\ \hline
Positive  										& \textbf{No changes}  		& \textbf{Small increase} 	& \textbf{Big increase} 	\\ \hline
\end {tabular}
\label{tab:rulebase}
\end{table}

\section{EXPERIMENTAL SETUP}
\label{sec:setup}

The experimental platform used in this work is Parrot Bebop 2 quadcopter UAV. This UAV is controlled via a Wi-Fi connection and the robot operating system (ROS) is used to communicate with UAV. The motion capture system provides the UAV's real-time position at $240 \si{Hz}$. This position is fed into the ground station computer (CPU: $2.6 \si{GHz}$, $64 \si{bit}$, quad-core; GPU: $4 \si{GB}$; RAM: $16 \si{GB}$ DDR4) where the algorithms are executed. Once the control signal is computed, it is sent to the UAV at $100 \si{Hz}$ rate.

For the attitude/velocity tracking, the onboard nonlinear geometric controller on $\mathsf{SE}(3)$ is used \cite{Lee2013AJC}. The attitude controller is responsible for mapping the high-level control inputs, i.e., $\begin{bmatrix} \theta^*_k & \phi^*_k & w^*_k \end{bmatrix}$, to the low-level control commands, i.e., $\mathbf{u}_k$ in (\ref{eq:inputs}).

\subsection{Deep Neural Network Structure}

Three feed-forward DNNs with hyperbolic tangent ($\tanh$) activation functions are used to learn the control mapping for each controlled axis: $x$, $y$ and $z$. The inputs to DNN for the $x$-axis are the errors and their time derivatives on the $x$-axis, $\{ e_{x, k}, e_{x, k - 1}, e_{x, k - 2}, \dot{e}_{x, k}, \dot{e}_{x, k - 1}, \dot{e}_{x, k - 2}\}$, and the output is the desired pitch angle, $\{ \theta^*_k \}$. Similarly, the inputs to DNN for the $y$-axis are the errors and their time derivatives on the $y$-axis, $\{ e_{y, k}, e_{y, k - 1}, e_{y, k - 2}, \dot{e}_{y, k}, \dot{e}_{y, k - 1}, \dot{e}_{y, k - 2} \}$, and the output is the desired roll angle, $\{ \phi^*_k \}$. Finally, the inputs to DNN for the $z$-axis are the errors and their time derivatives on the $z$-axis, $\{ e_{z, k}, e_{z, k - 1}, e_{z, k - 2}, \dot{e}_{z, k}, \dot{e}_{z, k - 1}, \dot{e}_{z, k - 2} \}$, and the output is the desired vertical velocity, $\{ w^*_k \}$.

\begin{remark}
Both DNN controllers with and without online learning consist of three parallel sub-networks for $x$, $y$ and $z$ axes.
\end{remark}

In our case, after some heuristic analysis and experimental trials, the architecture of each network is chosen to consist of $6$ input neurons ($n_I = 6$), $6$ scaling neurons, $2$ fully connected hidden layers ($n_L = 2$) with $6$ neurons in each layer ($n_H = 6$), $1$ unscaling neuron and $1$ output neuron ($n_O = 1$). From the asymptotic analysis, the runtime complexity for the forward-propagation is $\mathrm{O}(n_L \cdot n_H^3 + n_L \cdot n_H) \equiv \mathrm{O}(n_L \cdot n_H^3)$. While the runtime complexity for the back-propagation is $\mathrm{O}(n_{QN} \cdot n_L \cdot n_H^4 + n_L \cdot n_H^3) \equiv \mathrm{O}(n_{QN} \cdot n_L \cdot n_H^4)$, where $n_{QN}$ is the number of iterations in the quasi-Newton method. Moreover, the runtime complexity for the fuzzy mapping in (\ref{eq:fm}) is constant w.r.t. the architecture of the network, i.e., $\mathrm{O}(1)$. The dominant operation in DNN$_0$ is the forward-propagation; therefore, the runtime complexity of DNN$_0$ is polynomial. However, DNN with online learning involves both forward-propagation and back-propagation; therefore, the runtime complexity of DNN is also polynomial but asymptotic to $\mathrm{O}(n_{QN} \cdot n_L \cdot n_H^4)$. Therefore, the proposed architecture was chosen as a compromise between the learning capability of the neural network and the update time through the back-propagation.


The error type is an important term in the loss index, and, in our case, it is chosen as the normalized squared error. The initialization algorithm is used to bring the neural network to a stable region of the loss function, and, in our case, it is selected as the random search. The training algorithm is the core part of the training, and, in our case, the quasi-Newton method is the most suitable choice for both offline and online training.

\begin{algorithm}[!t]
	\SetAlgoLined
	\DontPrintSemicolon
	\KwIn{Pre-trained DNN$_0$}
	\KwOut{Trained DNN}
	\KwResult{Learns and controls the system online}
	\Begin{
		DNN $\gets$ DNN$_0$\;			
		\Repeat{landing}{
			Get $\mathbf{y}_k$, $\mathbf{y}^*_k$, $\dot{\mathbf{y}}_k$ and $\dot{\mathbf{y}}^*_k$\;
			$\mathbf{e}_k \gets \mathbf{y}^*_k - \mathbf{y}_k$ by using (\ref{eq:error})\;
			$\dot{\mathbf{e}}_k \gets \dot{\mathbf{y}}^*_k - \dot{\mathbf{y}}_k$ by using (\ref{eq:d_error})\;
			$\Delta \mathbf{u}_k \gets$ FLS$(\mathbf{e}_k, \dot{\mathbf{e}}_k)$ by using (\ref{eq:fm})\;
			Calculate $\mathbf{u}_k$ by forward-propagation\;
			Update by back-propagation $<\{ \mathbf{e}_k, \dot{\mathbf{e}}_k \}, \{ \mathbf{u}_k + \Delta \mathbf{u}_k \}>$\;
		}
	}
	\label{alg:online}
	\caption{Online post-training of DNN.}
\end{algorithm}

\subsection{Data Collection}

To prepare the training samples of the flight data, the system was controlled by a conventional controller alone, while the position errors and their time derivatives were collected as training inputs, and the control signal was saved as the labelled output. By using PID controller, $100'000$ instances have been collected in the training dataset for each axis. This dataset is large enough for our application, however, the proposed method does not have any limitations on the dataset size. The training data include slow circular and eight-shaped trajectories on $xy$-, $xz$- and $yz$-planes with the reference speed of $1\si{m/s}$.

\begin{figure*}[!b]
\centering
\subfloat[3D view for the tracking of the slow circular trajectory.]{\includegraphics[width=0.73\columnwidth]{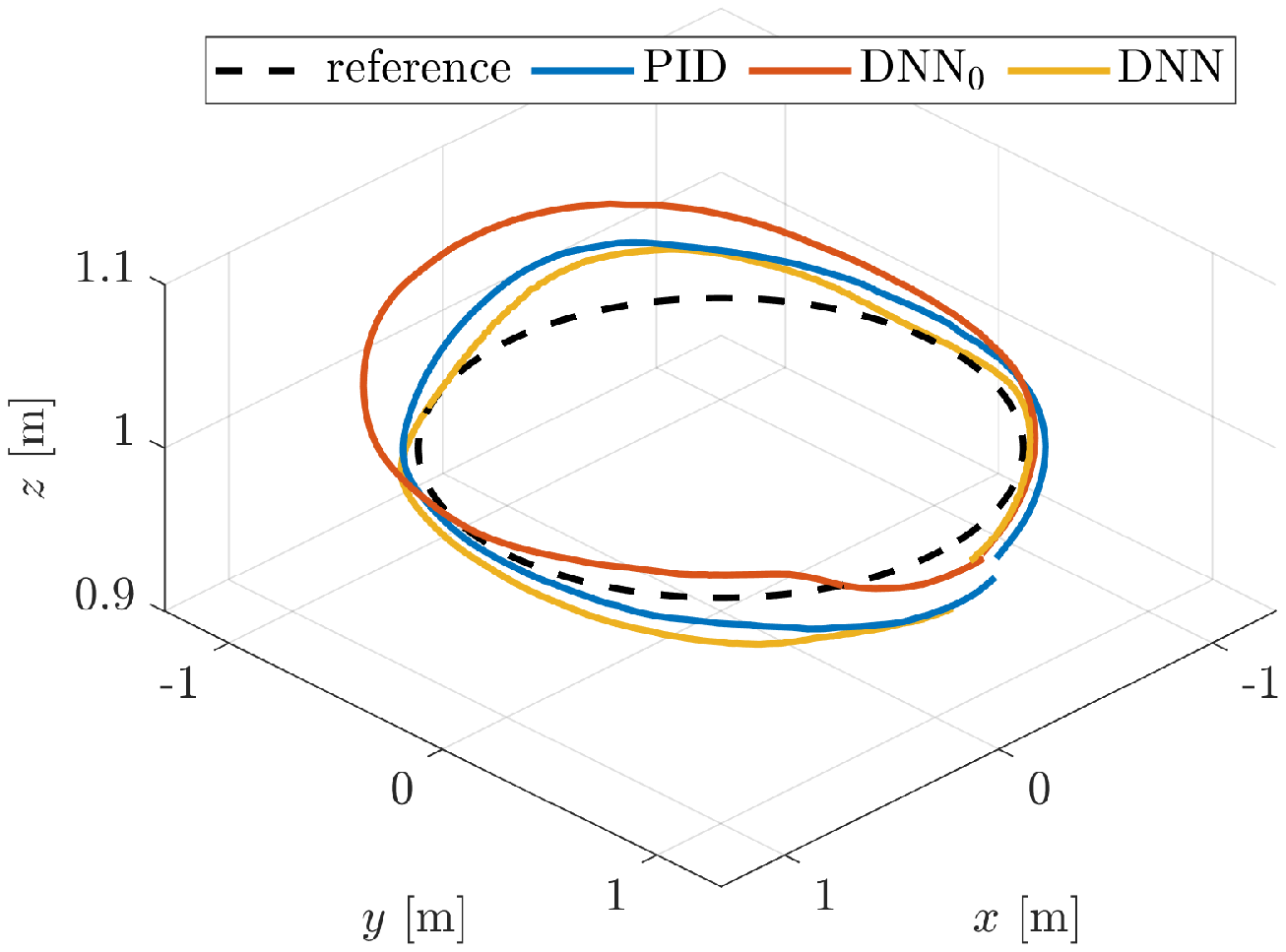}%
\label{fig:circle_3d}} \hfil
\subfloat[Projection of the slow circular trajectory tracking on $x$, $y$ and $z$ axes.]{\includegraphics[width=0.62\columnwidth]{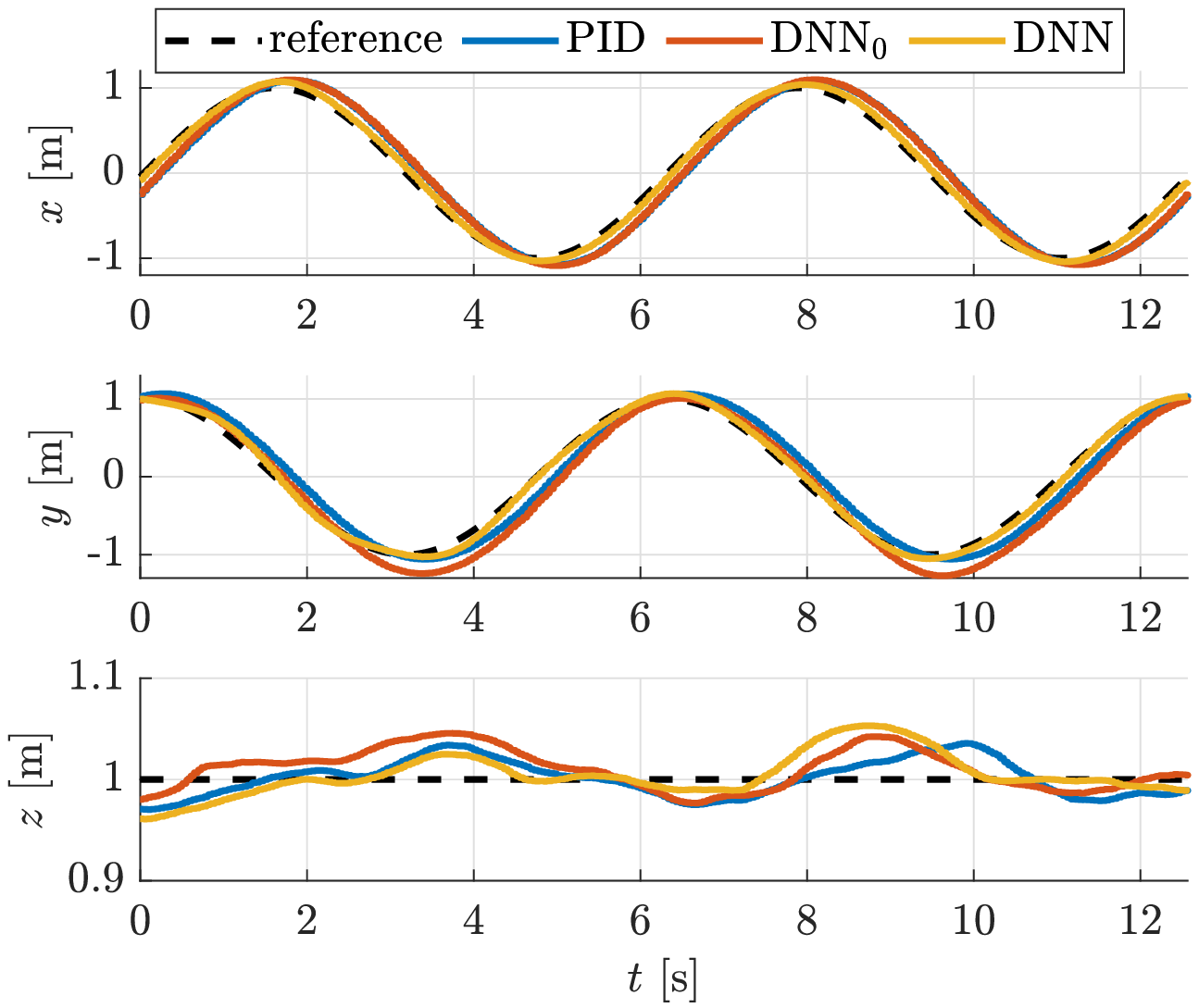}%
\label{fig:circle_xyz}} \hfil
\subfloat[Euclidean error for the slow circular trajectory tracking.]{\includegraphics[width=0.65\columnwidth]{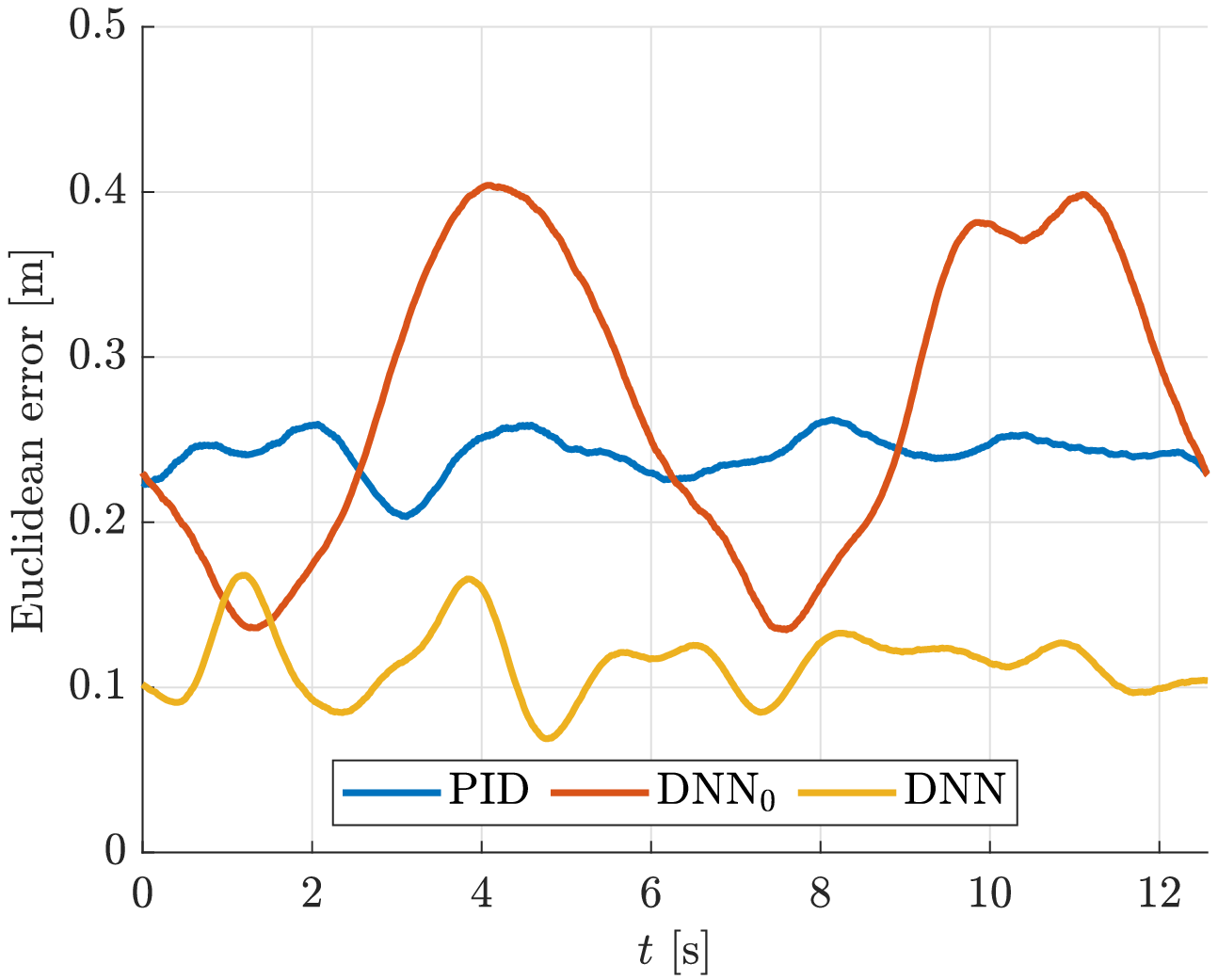}%
\label{fig:circle_error}} \\
\subfloat[3D view for the tracking of the fast circular trajectory.]{\includegraphics[width=0.73\columnwidth]{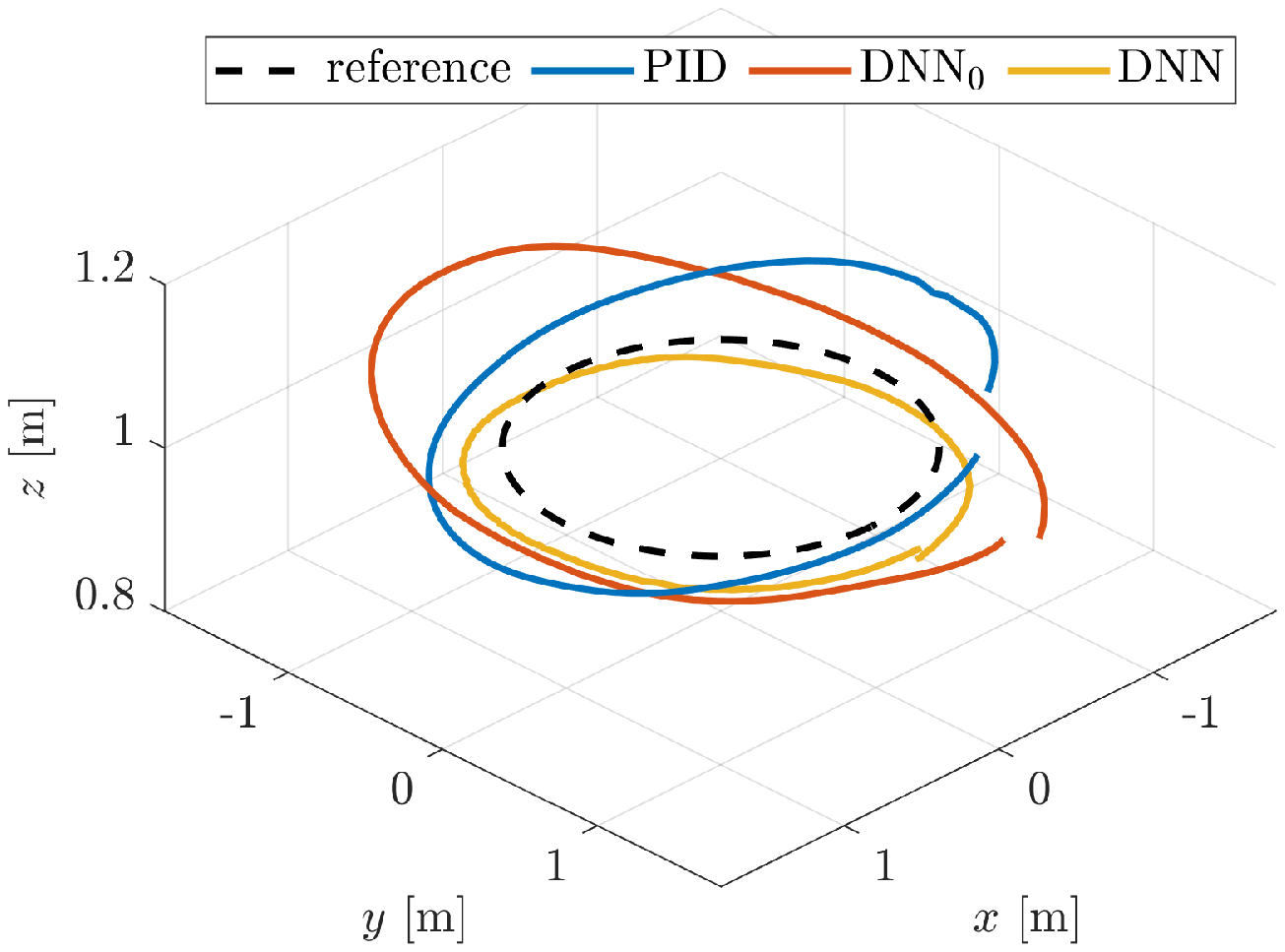}%
\label{fig:fast_circle_3d}} \hfil
\subfloat[Projection of the fast circular trajectory tracking on $x$, $y$ and $z$ axes.]{\includegraphics[width=0.62\columnwidth]{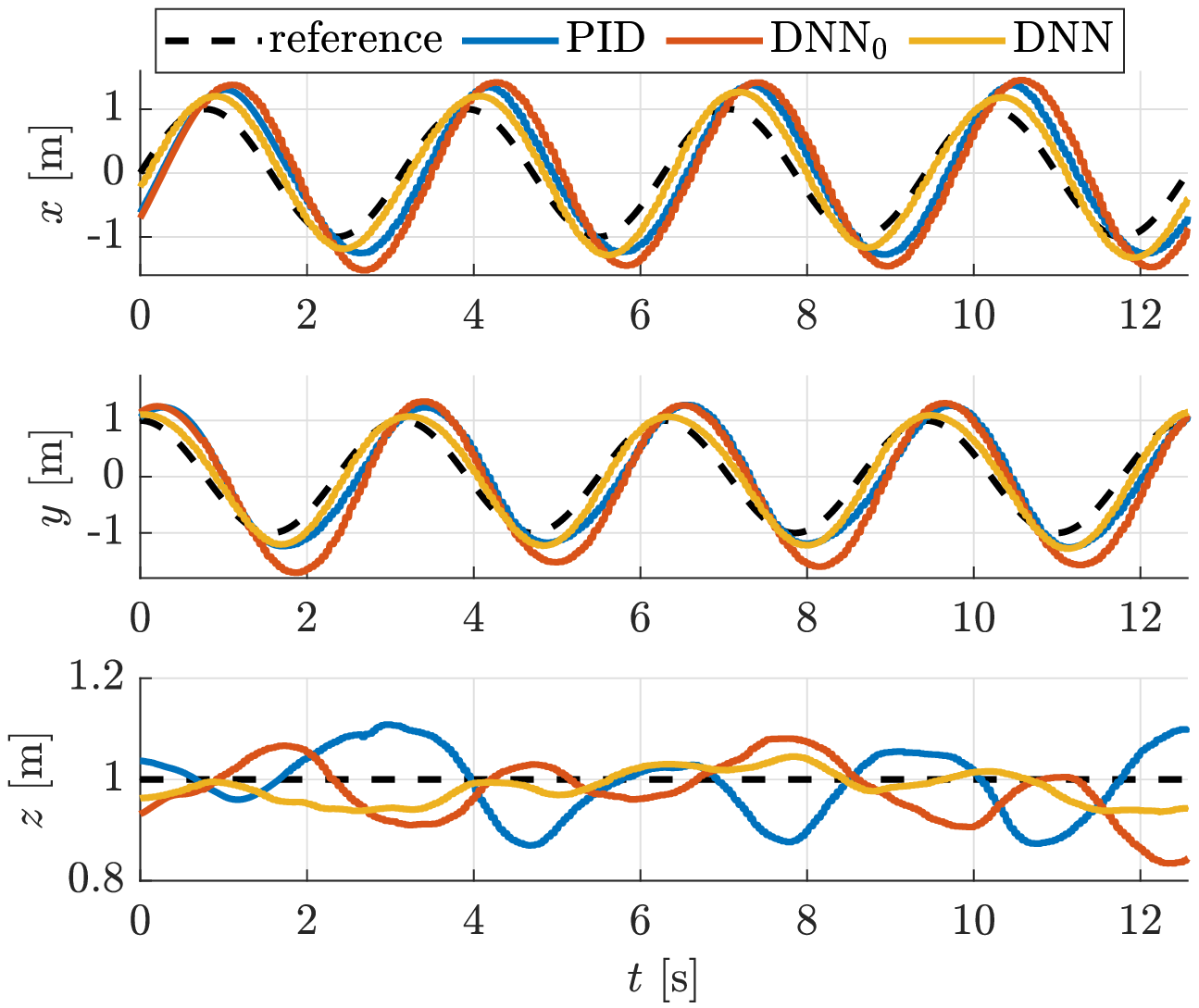}%
\label{fig:fast_circle_xyz}} \hfil
\subfloat[Euclidean error for the fact circular trajectory tracking.]{\includegraphics[width=0.65\columnwidth]{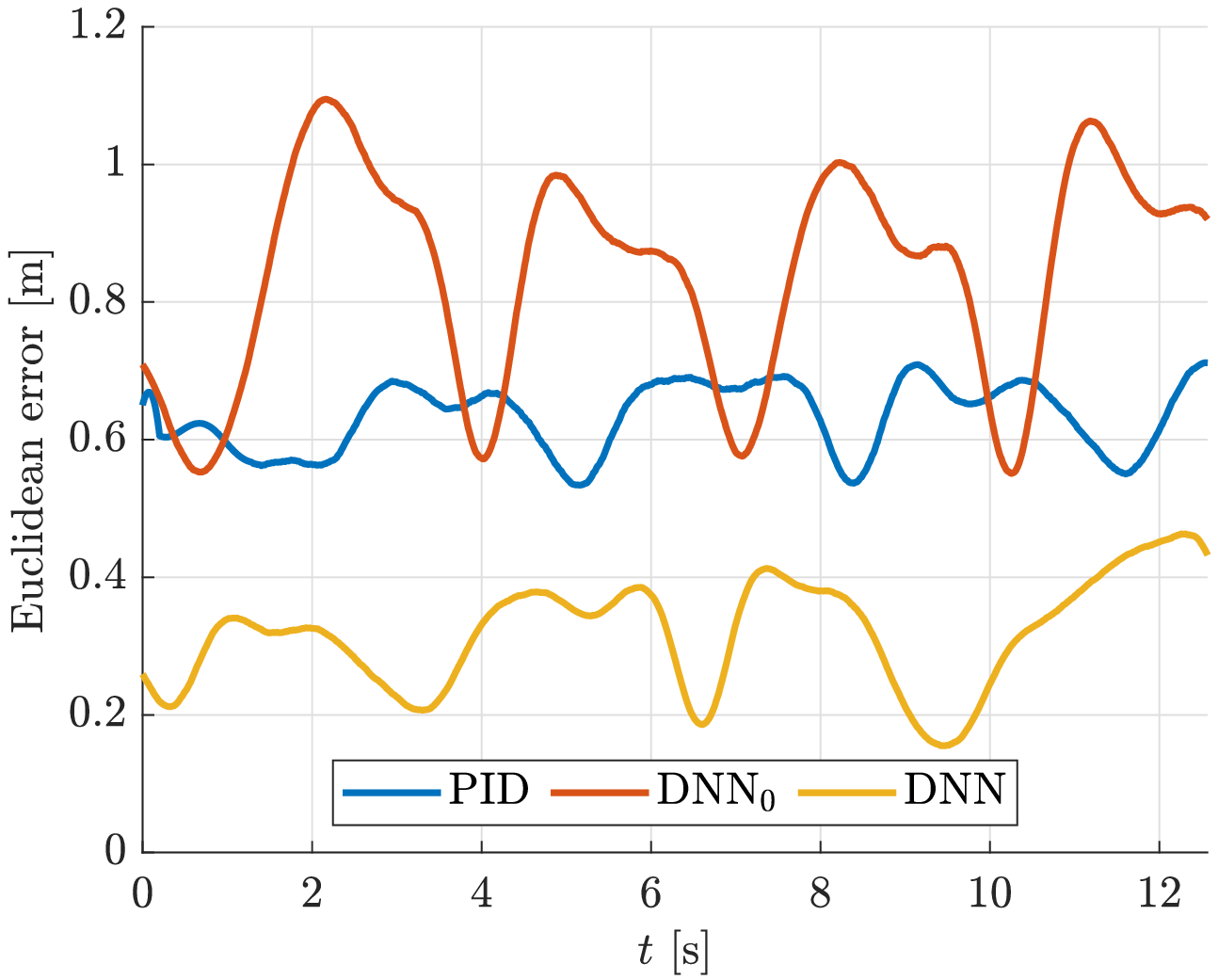}%
\label{fig:fast_circle_error}} \\
\subfloat[3D view for the tracking of the square-shaped trajectory.]{\includegraphics[width=0.73\columnwidth]{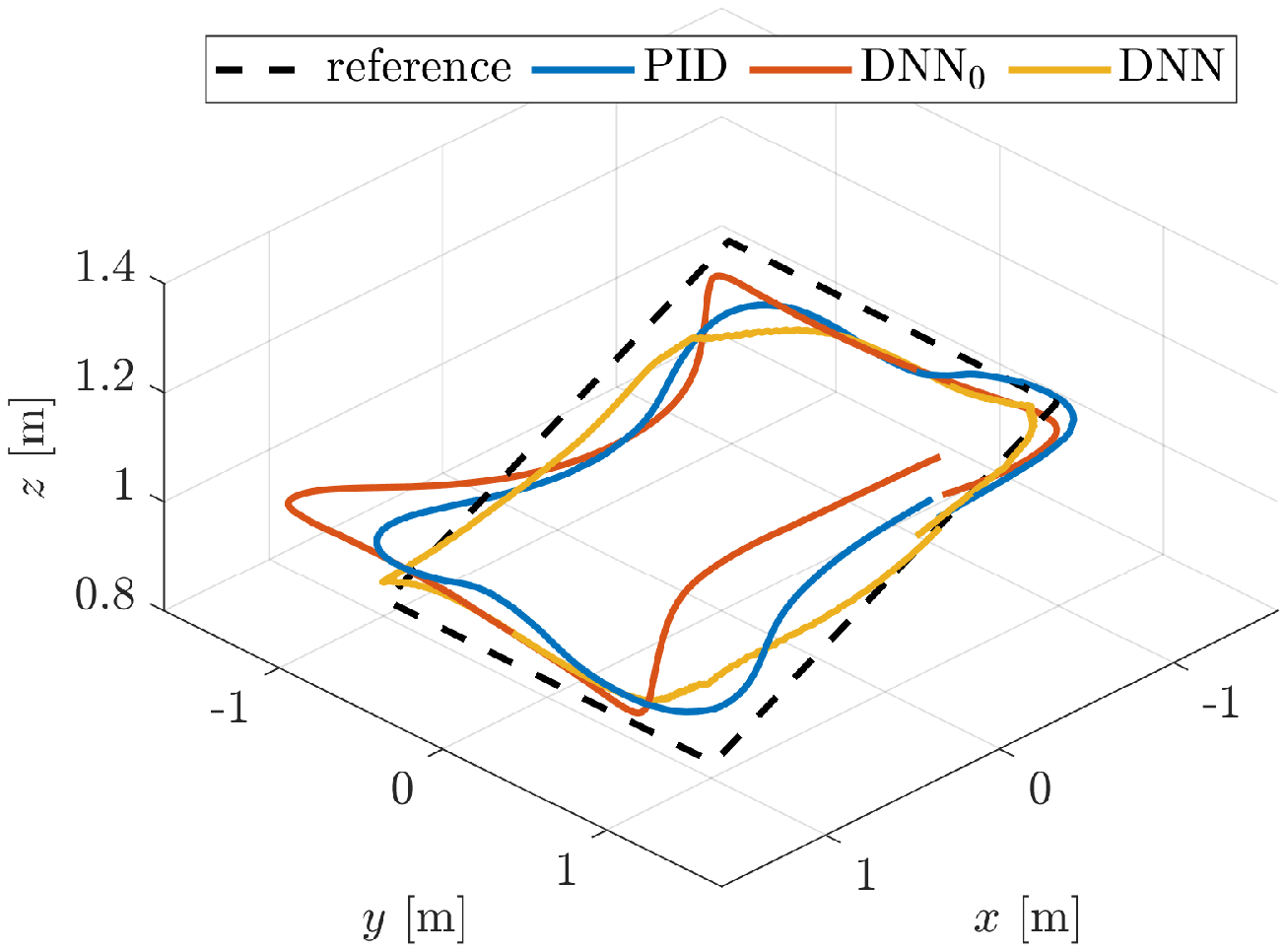}%
\label{fig:square_3d}} \hfil
\subfloat[Projection of the square-shaped trajectory tracking on $x$, $y$ and $z$ axes.]{\includegraphics[width=0.62\columnwidth]{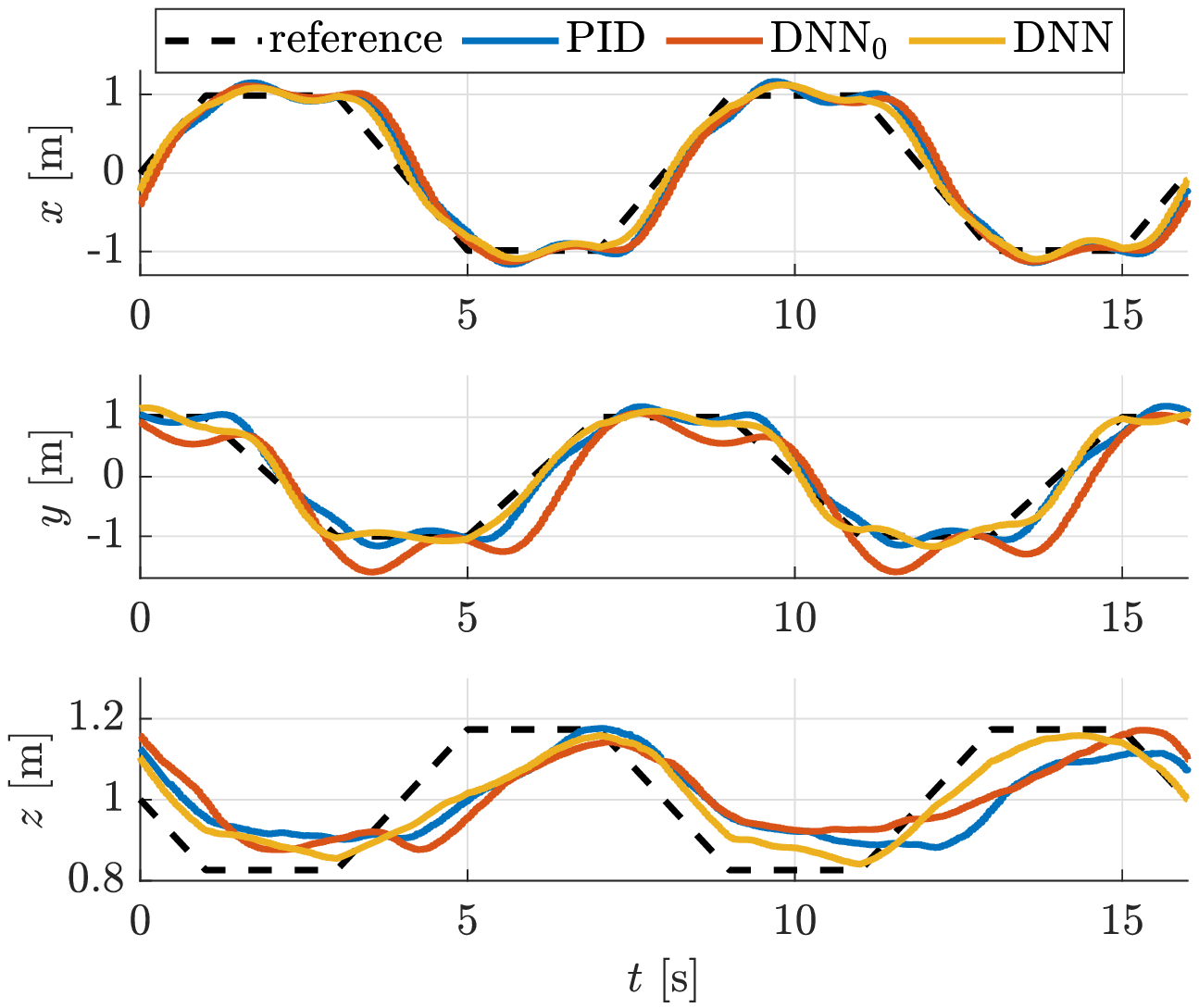}%
\label{fig:square_xyz}} \hfil
\subfloat[Euclidean error for the square-shaped trajectory tracking.]{\includegraphics[width=0.65\columnwidth]{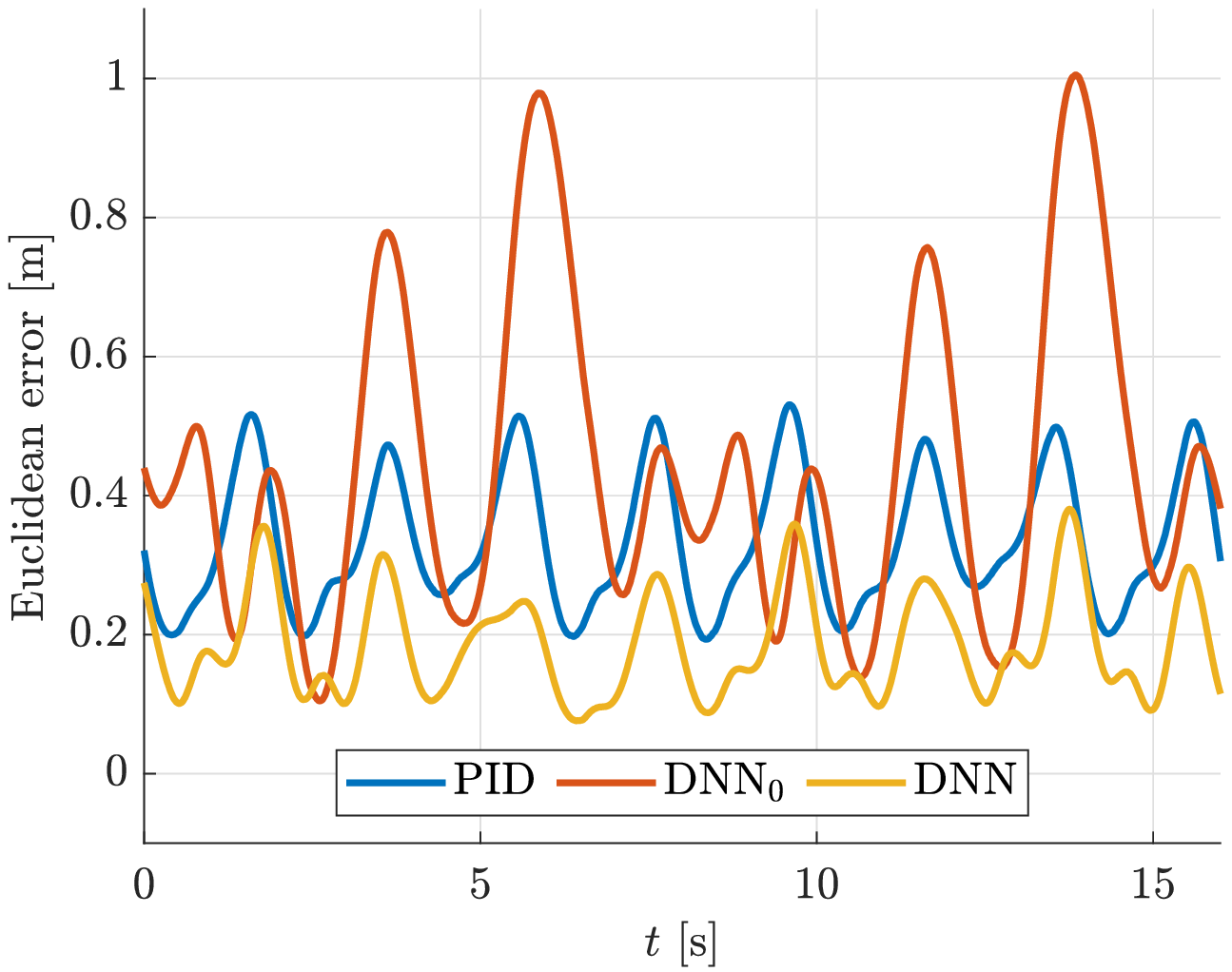}%
\label{fig:square_error}}
\caption{Experimental results for different controllers (PID, DNN$_0$ and DNN) on three trajectories (slow circular, fast circular and square-shaped). The slow circular trajectory has been used for the offline pre-training of DNN$_0$, while fast circular and square-shaped trajectories have not been used during the pre-training. It is possible to observe that DNN controller with online training is able to learn the system dynamics and improve the tracking performances on all tested trajectories.}
\label{fig:experiments}
\end{figure*}

\section{EXPERIMENTAL RESULTS}
\label{sec:results}

In order to validate the capabilities of the proposed controller in Section~\ref{sec:dnn}, the trajectory following problem of a quadcopter UAV is considered. The proposed control architecture and its training process are depicted in Fig.~\ref{fig:architecture}. Three different types of trajectories have been tested: slow circular, fast circular and square-shaped. In order to show the efficiency and efficacy of the DNN-based controller, it is compared with a well-tuned PID controller (used during the offline pre-training) and DNN controller without online training, DNN$_0$.

The first study case is the tracking of the slow circular trajectory with radius $1\si{m}$ at $1\si{m/s}$ which has been used during the pre-training phase. Fig.~\ref{fig:circle_3d} shows the results of the 3D trajectory tracking for the first case. The projections on $x$, $y$ and $z$ axes of this portion of the trajectory are shown on Fig.~\ref{fig:circle_xyz}. The evolution of the Euclidean error for the tested controllers is illustrated in Fig.~\ref{fig:circle_error}.
The second study case is the tracking of the fast circular trajectory with radius $1\si{m}$ at $2\si{m/s}$ which has not been used during the pre-training phase. Fig.~\ref{fig:fast_circle_3d} shows the results of the 3D trajectory tracking of the second case. The projections on $x$, $y$ and $z$ axes of this portion of the trajectory are shown on Fig.~\ref{fig:fast_circle_xyz}. The evolution of the Euclidean error for the tested controllers is illustrated in Fig.~\ref{fig:fast_circle_error}.
The third study case is the tracking of the square-shaped trajectory with side length $2\si{m}$ at $1\si{m/s}$ which also has not been used during the pre-training phase. Fig.~\ref{fig:square_3d} shows the results of the 3D trajectory tracking of the third case. The projections on $x$, $y$ and $z$ axes of this portion of the trajectory are shown on Fig.~\ref{fig:square_xyz}. The evolution of the Euclidean error for the tested controllers is illustrated in Fig.~\ref{fig:square_error}.

\subsection{Discussion}

A sample of experimental results for three controllers (PID, DNN$_0$ and DNN) on three trajectories (slow circular, fast circular and square-shaped) are illustrated on Figs.~\ref{fig:experiments}. It is possible to observe that DNN controller with online training is able to learn the system dynamics and decrease the tracking error over time on all tested trajectories. As visualized from Figs.~\ref{fig:circle_xyz},~\ref{fig:fast_circle_xyz} and ~\ref{fig:square_xyz}, DNN has faster responses, since it is able to estimate the desired control signal in (\ref{eq:system_inversion}) and predict the evolution of the system dynamics. It has to be emphasised that online DNN evolves from pre-trained DNN$_0$ during the learning process. Moreover, as expected, DNN$_0$ without online learning has poor performances on the trajectories which have not been used for its training.

\begin{figure}[!b]
\centering
\includegraphics[width=0.9\columnwidth]{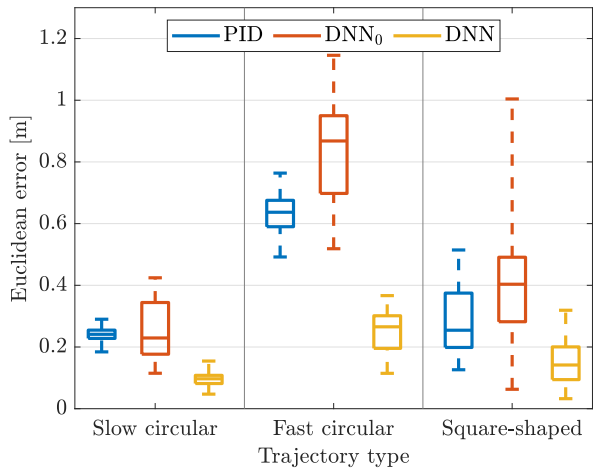}
\caption{Box-plot of the tracking performances of different controllers on three trajectories. For each trajectory, the experiments are repeated five times under the same conditions. It is possible to observe that in average the DNN controller with online learning outperforms other controllers on the tested trajectories.}
\label{fig:box_plot}
\end{figure}

For a statistical analysis of control performances, the experiments are repeated five times for each trajectory-controller combination under the same conditions. Fig.~\ref{fig:box_plot} presents a box-plot to compare the tracking performances of three different controllers on three tested trajectories. It is possible to observe that on average DNN controller with online learning outperforms other controllers on the tested trajectories. In addition, the maximum absolute error is also lower for the online DNN-based controller, even for previously unseen trajectory. Finally, the variance of the error is similar for PID and DNN with online learning controllers.

As can be seen from Table~\ref{tab:mae}, the DNN-based controller with online learning outperforms both PID and DNN$_0$ for all tested trajectory in terms of mean absolute error (MAE). Averaged results from numerous experiments depict that the overall improvement of $60\%$, $61\%$ and $46\%$ in MSE is achieved as compared to a well-tuned PID controller for slow circular, fast circular and square-based trajectories, respectively. While this ratio goes up to $62\%$, $70\%$ and $64\%$ when compared with pre-trained DNN$_0$ for the same trajectories.

\begin{table}[!t]
\caption{Comparison of different controllers in terms of MAE [$\si{m}]$.}
\centering
\begin{tabular}{|l||c|c|c|} \hline
\textbf{Trajectory} 	& \textbf{PID} 	& \textbf{DNN$_0$} 	& \textbf{DNN} 				\\ \hline \hline
Slow circle 			& $0.241$ 		& $0.254$ 			& \color{blue}{$0.097$} 	\\ \hline
Fast circle 			& $0.632$ 		& $0.833$ 			& \color{blue}{$0.250$} 	\\ \hline
Square-shaped 			& $0.284$ 		& $0.431$ 			& \color{blue}{$0.154$} 	\\ \hline
\end{tabular}
\label{tab:mae}
\end{table}


Though the online DNN-based controllers can learn promptly how to control the system, the computing time is still the main drawback of this controller with online back-propagation. The computing time is polynomially proportional to the number of hidden layers and the number of neurons in each layer. Therefore, deeper is the network, more complex functions it can learn but more computational power it requires. The average experimental computation time for DNN with online back-propagation is around $5.4\si{ms}$, while for PID and DNN$_0$ without online learning this time is only $8\si{\mu s}$ and $16\si{\mu s}$, respectively. However, $5.4\si{ms}$ is still an acceptable time for real-time applications, which allows the controller to run at almost $200\si{Hz}$.

\section{CONCLUSIONS}
\label{sec:conclusions}

In this work, we have presented a novel approach for a high-level control of UAV that improves online the trajectory tracking performances by using deep learning and expert knowledge. The learning is subdivided into two phases: offline pre-training and online training. During the offline learning phase, a conventional controller performs a set of trajectories and the batch of training samples is collected. Then, DNN-based controller, DNN$_0$, is pre-trained on the collected data samples. However, DNN$_0$ cannot adapt to the new flying conditions unseen during the pre-training; therefore, the online training is required. During the online learning phase, DNN controls the system and adapts the control input to improve the tracking performance. The expert knowledge encoded into the rule-base, thanks to the fuzzy mapping, provides the adaptation information to DNN allowing the real-time learning. Once DNNs are trained during the flight on UAV, the experimental results show that the proposed approach improves the performance by around 50\%. We believe that the results of this study will open the doors to a wider use of DNN-based controllers with online training in real-world control applications as the proposed structure is suitable to deploy in real-time control systems.

In the future, we will test the DNN-based controller for the aerial transportation where the system dynamics change drastically. In addition, we will extensively analyse the parameters and architecture of DNN and their performances. Moreover, the analytical stability proof of the proposed approach will be provided.

\section*{ACKNOWLEDGMENT}

This research was partially supported by the Singapore Ministry of Education (RG185/17).

\bibliographystyle{IEEEtran}
\bibliography{References}

\end{document}